\def\year{2022}\relax
\documentclass[letterpaper]{article} 
\usepackage{aaai22}  
\usepackage{times}  
\usepackage{helvet}  
\usepackage{courier}  
\usepackage[hyphens]{url}  
\usepackage{graphicx} 
\usepackage{natbib}  
\usepackage{caption} 
\DeclareCaptionStyle{ruled}{labelfont=normalfont,labelsep=colon,strut=off} 
\usepackage{algorithm}
\usepackage{algorithmic}
\usepackage{amsmath,amssymb,amsfonts}
\usepackage{newfloat}
\usepackage{listings}
\floatstyle{ruled}
\newfloat{listing}{tb}{lst}{}
\floatname{listing}{Listing}
\title{Learning from the Dark: Boosting Graph Convolutional Neural Networks with Diverse Negative Samples}
\author{
    Wei Duan\textsuperscript{\rm 1}, Junyu Xuan\textsuperscript{\rm 1}, Maoying Qiao \textsuperscript{\rm 2}, Jie Lu\textsuperscript{\rm 1}
}
\affiliations{
    \textsuperscript{\rm 1} The Australian Artificial Intelligence Institute (AAII), University of Technology Sydney\\
    \textsuperscript{\rm 2} Australian Catholic University\\

    Wei.Duan@student.uts.edu.au, Junyu.Xuan@uts.edu.au, maoying.qiao@acu.edu.au, Jie.Lu@uts.edu.au
}

\usepackage{bibentry}

\begin{document}

\maketitle

\begin{abstract}
Graph Convolutional Neural Networks (GCNs) has been generally accepted to be an effective tool for node representations learning. An interesting way to understand GCNs is to think of them as a message passing mechanism where each node updates its representation by accepting information from its neighbours (also known as positive samples). However, beyond these neighbouring nodes, graphs have a large, dark, all-but forgotten world in which we find the non-neighbouring nodes (negative samples).
In this paper, we show that this great dark world holds a substantial amount of information that might be useful for representation learning. Most specifically, it can provide negative information about the node representations. Our overall idea is to select appropriate negative samples for each node and incorporate the negative information contained in these samples into the representation updates. Moreover, we show that the process of selecting the negative samples is not trivial. Our theme therefore begins by describing the criteria for a good negative sample, followed by a determinantal point process algorithm for efficiently obtaining such samples. A GCN, boosted by diverse negative samples, then jointly considers the positive and negative information when passing messages. Experimental evaluations show that this idea not only improves the overall performance of standard representation learning but also significantly alleviates over-smoothing problems.
\end{abstract}

\section{Introduction}

\begin{figure}[!t]
\centering
\includegraphics[width=0.9\columnwidth]{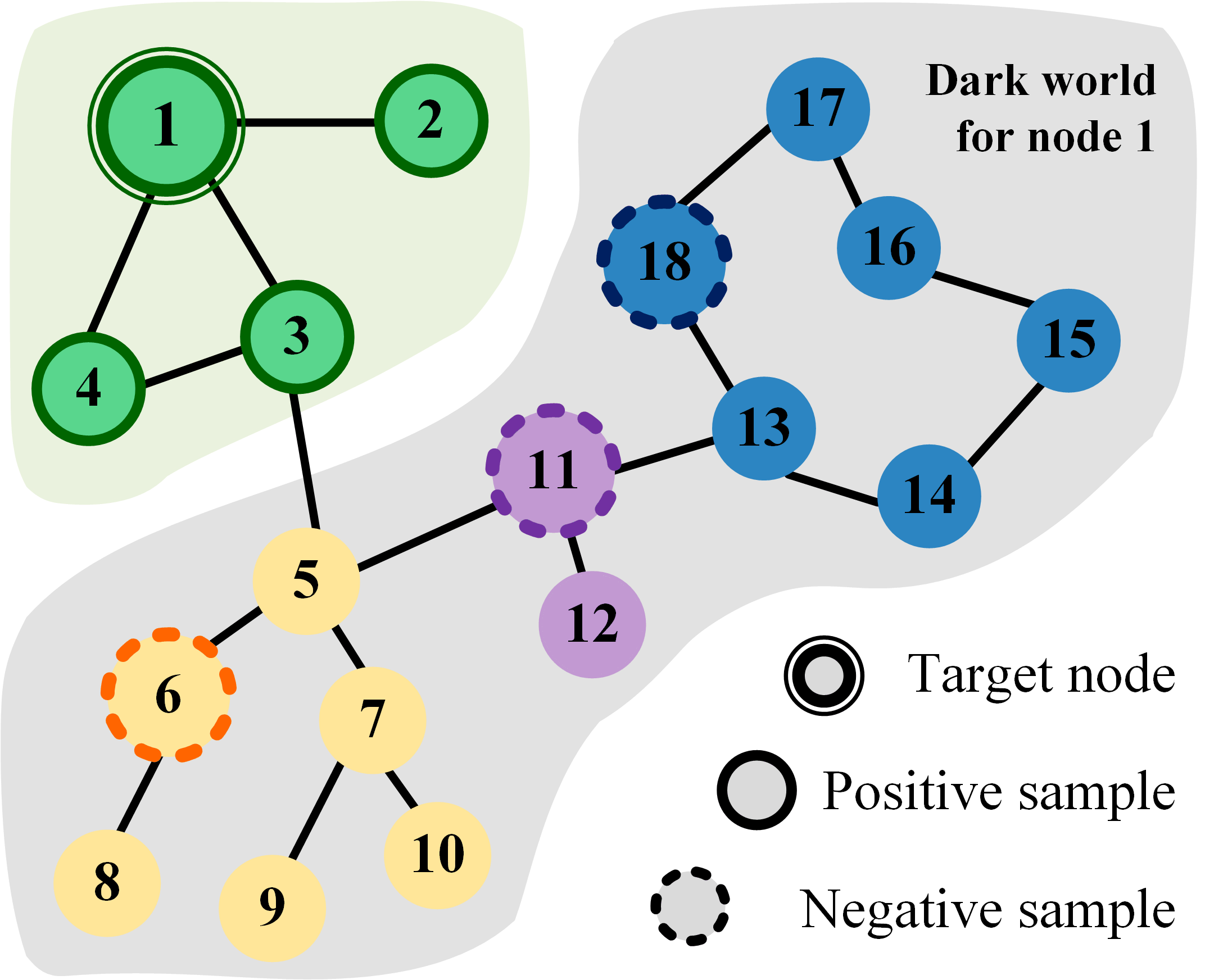} 
\caption{Illustration of the motivation of this work, including the dark world (gray shadow), different semantic clusters (green, yellow, purple, blue nodes), positive samples (nodes 2, 3, 4) and selected diverse negative samples (nodes 6, 11, and 18) of a given node (node 1).}
\label{fig:illustration}
\end{figure}

Graphs are powerful structures for modelling specific kinds of data such as molecules, social networks, citation networks, traffic networks, etc. \cite{chakrabarti2006graph}. However, the representation power of graphs is not a free lunch; it brings with it issues of incompatibility with some of the strongest and most popular deep learning algorithms, as these can often only handle regular data structures like vectors or arrays. Hence, to lend learning power to these great tools of representation, graph neural networks (GNNs) have emerged as a congruent deep learning architecture \cite{wu2020comprehensive}. Such has been their success that, today, there are many different GNN variants, each adapted for a specific task – for instance, graph sequence neural networks \cite{DBLP:journals/corr/LiTBZ15}, graph convolutional neural networks \cite{DBLP:conf/iclr/KipfW17}, and spatio-temporal graph convolutional networks \cite{DBLP:conf/ijcai/YuYZ18}.

Among all the varieties of GNNs, graph convolutional neural networks (GCN) \cite{DBLP:conf/iclr/KipfW17} is a simple but representative and salient one, which introduces the concept of convolution to GNNs that means to share weights for nodes within a layer. An easy and intuitive way to understand GCNs is to think of them as a message passing mechanism \cite{geerts2021let}  where each node accepts information from its neighbouring nodes to update its representation. The basic idea is that the representations of nodes with edge between them should be positively correlated. Hence, the neighboring nodes are also named as positive samples\footnote{Hereafter we refer to neighbouring nodes and positive samples interchangeably.}. This message-passing mechanism is highly effective in many scenarios, but it does lead to an annoying over-smoothing problem where the node representations become more and more similar as the number of layers of the graph neural network increases. This is hardly surprising when each node only updates its representation according to its neighbours. Yet, beyond these observed edges, there is a dark world that could provide diverse and useful information to the representation updates and help to overcome the over-smoothing problem at the same time, as illustrated in Fig. \ref{fig:illustration}. The reasoning is this: two nodes without edge between them should have different representations, so we can understand this as a negative link. However, in contrast to the commonly used positive links, these negative links are rarely used in the existing various GCNs.

Selecting appropriate negative samples in a graph is not trivial. To our knowledge, only three studies have explored procedures to this end. The first is Kim and Oh \cite{kim2021find}, who propose the natural and easy approach of uniformly randomly selecting some negative samples from non-neighbouring nodes. However, as we all know, a large graph is normally has the small-world property \cite{watts1998collective}  which means the graph will tend to contain some clusters. Moreover, nodes in the same cluster will tend to have similar representations while nodes within different clusters will tend to have different representations, as illustrated in Fig. \ref{fig:illustration}. Hence, under uniform random selection, the large clusters will overwhelm the smaller ones, and the final converged node representations will be short on information contained in the small clusters. Of the other two methods, one is based on Monte Carlo chains \cite{yang2020understanding} and the other on  personalised PageRank \cite{ying2018graph}, and neither covers diverse information as well. Further, the selected negative samples should not be redundant; each of them should not be overlapping and hold distinct information. Hence, as a definition, a good negative sample should \textit{contribute negative information to the given node contrast to its positive samples and include as much information as possible to reflect the variety of the dark world}.

In this paper, we propose a graph convolutional neural network boosted by diverse negative samples selected through a determinant point process (DPP). DPP is special point process for defining a probability distribution on a set where more diverse subset has a higher probability \cite{hough2009zeros}. Our idea is to find diverse negative samples for a given node by firstly defining a DPP-based distribution on diverse subsets of all non-neighbouring nodes of this node, and then outputting a diverse subset from this distribution. The number of subsets is limited because the samples are mutually exclusive. However, when applying this algorithm to large graphs, the computational cost can be as high as $O(N^3)$, where $N$ is the number of non-neighbouring nodes. Therefore, we propose a method based on a depth-first search (DFS) that collects diverse negative samples sequentially along the search path for a node. The motivation is that a DFS path is able to go through all different clusters in the graph and, therefore, the local samples collected will form a good, diverse approximation of all the information in the dark world. Further, the computational cost is approximately  $O(\overline{\text{pa}} \cdot \overline{\text{deg}}^3)$ where $\overline{\text{pa}} \ll N$ is the path length (normally smaller than the diameter of the graph) and $\overline{\text{deg}} \ll N$ is the average degree of the graph. As an example, consider a Cora graph \cite{sen2008collective} with 3,327 nodes, $\overline{\text{pa}} = 19$, and $\overline{\text{deg}} = 3.9$. Thus, $O(N^3) = 3.6e10 \gg 1,127 = O(\overline{\text{pa}} \cdot \overline{\text{deg}}^3)$.

In evaluating these ideas, we conducted empirical experiments on different tasks, including node classification and over-smoothing tests. The results show that our approach produces superior results.

Thus, the contributions of this study include:
\begin{itemize}
\item A new negative sampling method based on a DPP that selects diverse negative samples to better represent the dark information;
\item A DFS-based negative sampling method that sequentially collects locally diverse negative samples along the DFS path to greatly improve computational efficiency;
\item We are the first to fuse negative samples into the graph convolution, yielding a new GCN boosted by diverse negative samples (D2GCN) to improve the quality of node representations and alleviate the over-smoothing problem.

\end{itemize}

\begin{table}[!t]
      \begin{tabular}{c|p{0.75\columnwidth}}
        \hline
         Symbol & Meaning \\
         \hline 
         $G$ & a graph \\
         $G_n$ & the node set of graph $G$\\
         $\mathcal{Y}$ & a ground set \\
         $Y$ & a node subset\\
         $k$ & the number of negative samples of a given node in a graph\\
         $\mathcal{N}(i)$ &  neighbours (positive samples) of node $i$ \\
         $\overline{\mathcal{N}}(i)$ & negative samples of node $i$ \\
         $x_i^{(l)}$ & the representation of node $i$ at layer $l$ \\
         $\text{deg}(i)$ & the degree of node $i$ \\
         $\mathbb{L}$ & the $\mathbb{L}$-ensemble of DPP\\
         $\lambda$ & the eigenvalues of $\mathbb{L}$ \\
         $e_k^{|\mathcal{Y}|}$ & the $k^{th}$ elementary symmetric polynomial on eigenvalues $\lambda_1,\lambda_2,\dots, \lambda_{|\mathcal{Y}|}$ of $\mathbb{L}$ \\
         $\boldsymbol{v}$  & the eigenvectors of the $\mathbb{L}$-ensemble\\
         $ V$ & the set $\boldsymbol{v}$ \\
         \hline
      \end{tabular}
  \caption{Key notations}
  \label{tab:notation}
 \end{table}

\section{Preliminaries}
\label{sec:preliminary-knowledge}

This section briefly introduces the basic concepts of graph convolutional neural networks, message passing, and determinantal point process.

\subsection{Graph Convolutional Neural Networks (GCNs)}

GCNs introduce traditional convolution from classical neural networks to graph neural networks. Like a message-passing mechanism, the representation of a node is updated using its neighbours’ representations through

\begin{equation}
	{x_i}^{(l)} =\sum_{j \in \mathcal{N}_i \cup \{i\}}\frac{1}{\sqrt{\text{deg}(i)}\cdot\sqrt{\text{deg}(j)}}\big(\Theta^{(l)} \cdot x_j^{(l-1)}\big)
\label{eq:gcn}	
\end{equation} 
where $x_i^{(l)}$ are the representation of node $i$ at layer $l$, $\mathcal{N}(i)$ is the neighbours of node $i$ (i.e., positive samples), $\text{deg}(i)$ is the degree of node $i$, and $\Theta^{(l)}$ is a feature transition matrix. The goal of this equation is to aggregate (sum) the weighted representations of all neighbours. Note that although we use GCNs to demonstrate our approach throughout the paper, the same ideas can be easily applied to other variants as well.

\subsection{Determinantal Point Processes (DPP)}

Given a ground set $\mathcal{Y}=\{1, 2, \dots, |\mathcal{Y}|\}$, a determinantal point process $\mathcal{P}$ is a probability measure on all possible subsets of $\mathcal{Y}$ with size $2^{|\mathcal{Y}|}$. For every $Y \subseteq \mathcal{Y}$, a DPP \cite{hough2009zeros} defined via an $\mathbb{L}$-ensemble is formulated as
\begin{align}
\mathcal{P}_{\mathbb{L}}(Y) = \frac{\det(\mathbb{L}_Y)}{\det(\mathbb{L}+I)}
\end{align}
where $\det(\cdot)$ denotes the determinant of a given matrix, $\mathbb{L}$ is a real and symmetric $|\mathcal{Y}| \times |\mathcal{Y}|$ matrix indexed by the elements of $\mathcal{Y}$, and $\det(\mathbb{L}+I)$ is a normalisation term that is constant once the ground dataset $\mathcal{Y}$ is fixed.
Given a Gram decomposition $\mathbb{L}_Y = \bar{\mathbb{L}}^T \bar{\mathbb{L}}$, a determinantal operator can be interpreted geometrically as
\begin{align} 
\mathcal{P}_{\mathbb{L}}(Y) \propto \det(\mathbb{L}_Y) = \text{vol}^2 \left(\{\bar{\mathbb{L}}_i\}_{i\in Y} \right)
\end{align}
where the right hand side is the squared volume of the parallelepiped spanned by the columns in $\bar{\mathbb{L}}$ corresponding to elements in $Y$. Intuitively, to get a parallelepiped of greater volume, the columns should be as repulsive as possible to each other. Hence, DPP assigns a higher probability to a subset of $Y$ whose elements span a greater volume.

One important variant of DPP is $k$-DPP \cite{kulesza2012determinantal}. $k$-DPP measures only $k$-sized subsets of $\mathcal{Y}$ rather than all of them including an empty subset. It is formally defined as
\begin{align}
\mathcal{P}_{\mathbb{L}}(Y) = \frac{\det(\mathbb{L}_Y)}{e_k^{|\mathcal{Y}|}} 
\label{eq:kdpp}
\end{align}
with the cardinality of the subset $Y$ being a fixed size $k$, i.e., $|Y|=k$.
$e_k^{|\mathcal{Y}|}$ is the $k^{th}$ elementary symmetric polynomial on eigenvalues $\lambda_1,\lambda_2,\dots, \lambda_{|\mathcal{Y}|}$ of $\mathbb{L}$, i.e., $e_k(\lambda_1,\lambda_2,\dots, \lambda_{|\mathcal{Y}|})$.

Both DPP and $k$-DPP can be used to sample a diverse subset, and both have been well studied in the machine learning area \cite{kulesza2012determinantal}. The popularity of DPP (and also of $k$-DPP) for modeling diversity is because of its great modelling power.

\section{Related Work} 
\label{sec:related-work}

\subsection{GCN and Its Variants}

Motivated by the achievements of convolutional neural networks (CNNs), many GNNs approaches have been actively studied to model graph data. These are classified into two types, spectral-based and spatial-based. \cite{DBLP:journals/corr/BrunaZSL13} first created a graph convolution based on spectral graph theory. Although their paper was conceptually important, it had major computational flaws, which prevented it from being a genuinely useful tool. Since then, a growing number of enhancements, extensions, and approximations of spectral-based GCNs have been made to overcome these flaws. Based on these, \cite{DBLP:conf/iclr/KipfW17} proposed GCNs, which is a localized first-order approximation of spectral graph convolutions as a generalised method for semi-supervised learning on graph-structured data. Their model acquires implicit representations, encodes the region graph structure and node attributes, and expands linearly in terms of the number of graph edges.

A representative work in a spatial-based way is GraphSAGE \cite{hamilton2017inductive}, which is a general framework for generating node embedding by sampling and aggregating features from neighbourhood of a node. In order to theoretically analyze the representational power of GNNs, \cite{DBLP:conf/iclr/XuHLJ19} formally characterised how expressive different GNNs variants are at learning to represent different graph structures based on the graph isomorphism test \cite{weisfeiler1968reduction}. They proposed that, since modern GNNs follow a neighbourhood aggregation strategy, the network at the $k$-th layer can be summarised formally in two steps AGGREGATE and COMBINE. In addition, they further proposed GINs, which can distinguish different graph structures and capture dependencies between graph structures to achieve better classification results \cite{DBLP:conf/iclr/XuHLJ19}. 

All the above GNNs can be considered to use positive sampling when generating new node feature vectors,  because the neighbour aggregation on $j \in \mathcal{N}(i)$  leads to a high correlation between the central node and its neighbours. Extremely few studies use negative sampling with a GNN. The only three works \cite{ying2018graph,kim2021find,yang2020understanding} that do use it do not satisfy the our criteria of good negative samples: good negative samples should include as many information of the dark world as possible and, at the same time, without much overlapping and redundant information. Moreover, the above approaches only use the results of negative sampling in the loss function, while the direct application of to convolution operations remaining unexplored.


\subsection{DPP and Its Applications}

Determinantal point processes (DPPs) \cite{hough2009zeros} are statistical models and provide probability measures over every configuration of subsets on data points. DPPs were first introduced to machine learning area by \cite{kulesza2012determinantal}, and they have since been extended to include closed-form normalisation, marginalisation \cite{kulesza2012determinantal}, sampling \cite{kang2013fast}, dual
representation, maximising a posterior (MAP) \cite{gillenwater2012near}) and parameter learning \cite{affandi2014learning,gillenwater2014expectation}, and its structural \cite{gillenwater2012discovering} and Markov \cite{affandi2012markov} variants to name just two. This repulsive characteristic has been successfully applied to prior modelling in a variety of scenarios, such as clustering \cite{kang2013fast}, 
inhibition in neural spiking data \cite{snoek2013determinantal}, sequential labelling \cite{qiao2015diversified}, document summarisation, video summarisation, tweet timeline generation, and so on. However, to the best of our knowledge, DPP has not been used for negative sampling with GNNs.

\section{Proposed Model}
\label{sec:our-proposed-model}

This section first introduces a method for obtaining good negative samples for a given node, including two negative sampling algorithms. We then integrate the algorithms with a GCN to obtain a new graph convolutional neural network boosted by diverse negative samples (D2GCN).

\subsection{DPP-Based Negative Sampling}

Given a node, we believe that its good negative samples should include as much information about the dark world as possible and, at the same time, without much overlap and redundancy. The repulsive property of DPP inspired us to use it to select negative samples. However, applying DPP to different scenarios is not trivial; modifications are required to make it work. Hence, for a given node $i$ in a graph $G$, we need to first define a $\mathbb{L}$-ensemble for our problem:
\begin{equation}
	\mathbb{L}_{G_n\setminus i}(j, j') = \exp \left(\cos \left(x_j, x_{j'} \right)-1 \right),
\label{eq:lensemble}	
\end{equation} 
where $G_n\setminus i$ denotes the node set of $G_n$ excluding node $i$, $j$ and $j'$ are two nodes within $G_n\setminus i$, and $\cos  (\cdot, \cdot)$ is the cosine similarity between two node representations $x$. The node feature $x$ is used here because, when defining the distance between nodes, nodes with similar information are expected to be further apart. $x_j$ is expected to encode the information of node $j$ in terms of both features and network structure. Feature information may dominate at the beginning but both types of information will be balanced after a number of training steps. The reason we chose cosine similarity is because the following prediction (label and link) is normally based on cosine similarity. We used only $x$ to simplify the explanation. However, it would be straightforward to include more information here, like the network distance between nodes $j$ and $j'$, the similarity with node $i$ or node degree of each node.

\begin{algorithm}[!t]
\caption{Diverse negative sampling}
\label{alg:dpp}
\textbf{Input}: A graph $G$ \\
\textbf{Output}: $\overline{\mathcal{N}}(i)$ for all $i \in G$
\begin{algorithmic}[1] 
\STATE Let $\overline{\mathcal{N}} = \bf{0}$.
\FOR{each node $i$}
\STATE Compute $\mathbb{L}_{G\setminus i}$ using Eq. (\ref{eq:lensemble});
\STATE Define a distribution on all possible subsets $\mathcal{P}_{\mathbb{L}}(Y_{G\setminus i})$ using Eq. (\ref{eq:py});
\STATE Obtain a sample of $\mathcal{P}_{\mathbb{L}}(Y_{G\setminus i})$ using Algorithm 8 in \cite{kulesza2012determinantal};
\STATE Save nodes in the sample as $\overline{\mathcal{N}}(i)$;
\ENDFOR
\STATE \textbf{return} $\overline{\mathcal{N}}$
\end{algorithmic}
\end{algorithm}

With this $\mathbb{L}$-ensemble, we can obtain a distribution on all possible subsets of all nodes in the graph except for $i$,
\begin{equation}
\mathcal{P}_{\mathbb{L}}(Y_{G_n\setminus i}) = \frac{\det(\mathbb{L}_{Y_{G_n\setminus i}})}{\det(\mathbb{L}_{G_n\setminus i}+I)}
\label{eq:py}
\end{equation} 
where $Y_{G_n\setminus i}$ is a set of all subsets of $G_n\setminus i$ and $\det (\cdot)$ is the matrix determinant operator. Comparing other ordinary probability distributions with same support, this distribution has a nice and unique property that the more diverse the subsets, the higher their probability values, the easier it is to obtain a diverse set from this distribution by sampling. To ensure the scale of negative samples is similar to the positive samples, we use $k$-DPP to fix the number of negative samples as $k=|\mathcal{N}_i|+1$. Here, we use a sampling method based on eigendecomposition \cite{hough2006determinantal,kulesza2012determinantal}. Eq. (\ref{eq:py}) can be rewritten as the $k$-DPP distribution in terms of the corresponding DPP
\begin{equation}
\mathcal{P}_{\mathbb{L}}^{k}(Y_{G_n\setminus i}) = \frac{1}{e_{k}^{|G_n\setminus i|}}\det(\mathbb{L}_{G_n\setminus i}+I)\mathcal{P}_{\mathbb{L}}(Y_{G_n\setminus i})
\label{eq:PLk}
\end{equation}
whenever $|Y_{G_n\setminus i}| = k $ and $e_{k}^{|G_n\setminus i|}$ denotes the $k$-th elementary symmetric polynomial. Following the \cite{kulesza2012determinantal}, Eq. (\ref{eq:PLk}) can be decomposed into elementary parts
\begin{equation}
\mathcal{P}_{\mathbb{L}}^{k}(Y_{G_n\setminus i}) = \frac{1}{e_{k}^{|G_n\setminus i|}}\sum_{|J|=k}\!\mathcal{P}^{V_{J}}(Y_{G_n\setminus i}) \!\prod_{m\in J}\! {\lambda}_m
\label{eq:PLkelement}
\end{equation}
where $ V_{Y_{G_n\setminus i}} $ denotes the set $\{\boldsymbol{v}_m\}_{m\in Y_{G_n \setminus i}}$ and $\boldsymbol{v}_m$ and ${\lambda}_m $ are the eigenvectors and eigenvalues of the $\mathbb{L}$-ensemble, respectively. Based on Eq. (\ref{eq:PLkelement}), the complete process of sampling from $k$-DPP is in Algorithm 8 in \cite{kulesza2012determinantal}.

Note that comparing with a sample, the mode of this distribution is a more rigorous output \cite{gillenwater2012near}, but it is usually with an unbearable complexity, so we use a sample rather than mode here. The experimental evaluation have shown that the sample has been able to achieve satisfactory results. Algorithm \ref{alg:dpp} shows a diverse negative sampling method based on DPP, but its computation cost is above the standard for a DPP on $G_n\setminus i$ is $O(|G_n\setminus i|^3)$ for each node. This totals an exorbitant $O(|G_n\setminus i|^4)$ for all nodes! Although Algorithm \ref{alg:dpp} is able to explore the whole dark world and find the best diverse negative samples, the large number of candidates makes it an impractical solution for even a moderately-sized graph. Hence, we propose the approximate heuristic method below.

\begin{figure}[!t]
\centering
\includegraphics[width=0.45\textwidth]{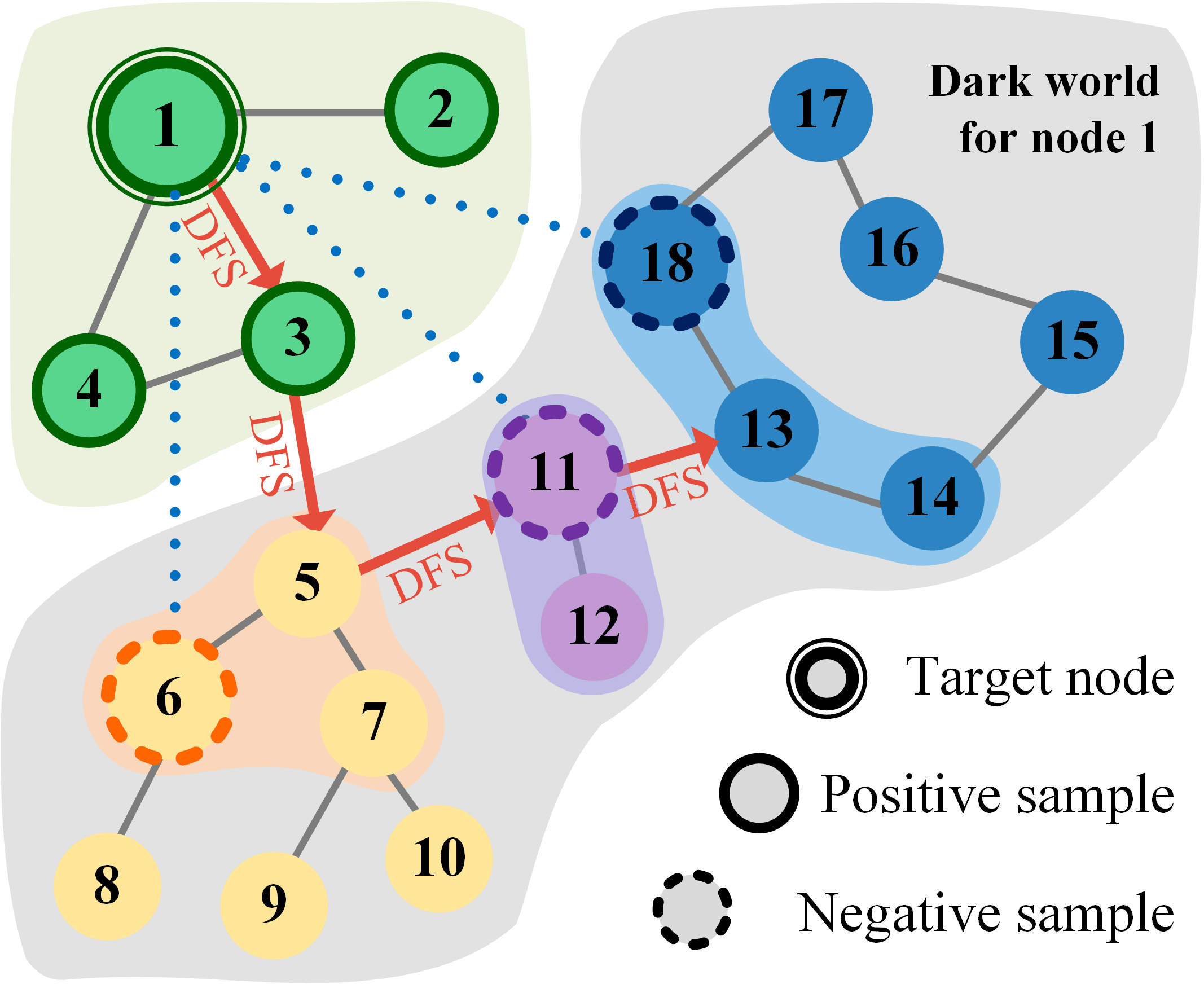} 
\caption{The concept of DPP-based negative sampling. The target node is Node 1. Nodes 2, 3 and 4 are positive samples. Nodes 5-18 are the dark world of Node 1. The 4-length DFS path of Node 1 is $\{3, 5, 11, 13\}$, where $\{5, 11, 13\}$ are the central nodes on the path in the dark world. With their first-order neighbouring nodes, they form the candidate set of DPPs, i.e.$\{5,6,7,11,12,13,14,18\}$. The selected negative samples from this set are 6, 11, and 18, which can be seen as virtual negative links to Node 1.}
\label{fig2}
\label{fig:ill2}
\end{figure}

Our belief is that the good negative samples are the ones with different semantics and complete knowledge of the whole graph, such as nodes 6, 11, and 18 in Fig. \ref{fig:ill2}. In this figure, we hope the selected negative samples could each belong to the different `cluster' and all `clusters' are represented by the negative samples. Hence, given a node $i$, we use a depth-first-search (DFS) method to build a fixed-length path from node $i$ to the other nodes. We then collect the first-order neighbours of the nodes on this path to form a candidate set. Finally, we select diverse negative samples from this candidate set using the same DPP idea as outlined above. This DFS-based method can be considered as using a local diverse negative samples to approximate the global diverse negative samples. The reason is that DFS has the capability of breaking through the `cluster' of the current node $i$ to reach the semantics of other `clusters'. So the first-order neighbours of nodes on the path are able to supply sufficient information from many `clusters' but at a much smaller size. Although only first-order neighbours are collected here, collecting a higher order may further increase approximation performance; however, it would also increase the computational cost. In terms of path length, we found that performance is sufficiently good when the path length is set to around a quarter of the graph’s diameter. The full procedure is summarised in Algorithm \ref{alg:dfs-dpp}.

Here, the overall compuational cost is $O(N \cdot |\overline{\text{pa}}| \cdot \overline{\text{deg}}^3)$ where $|\overline{\text{pa}}| \ll N$ is the path length (normally smaller than the diameter of the graph) and $\overline{\text{deg}} \ll N$ is the average degree of the graph. That is much smaller than $O(|G_n\setminus i|^4)$ 

\begin{algorithm}[!t]
\caption{DFS-based diverse negative sampling}
\label{alg:dfs-dpp}
\textbf{Input}: A graph $G$ \\
\textbf{Output}: $\overline{\mathcal{N}}(i)$ for all $i \in G$
\begin{algorithmic}[1] 
\STATE Let $\overline{\mathcal{N}} = \bf{0}$;
\FOR{each node $i$}
\STATE Build a DFS path $pa_i$ in $G\setminus i$;
\STATE Let $C_i=[]$;
\FOR{each node $j \in pa_i$}
    \STATE Collect first-order neighbors $\mathcal{N}(j)$ of $j$;
    \STATE Expand $C_i = [C_i, \mathcal{N}(j)]$;
\ENDFOR
\STATE Compute $\mathbb{L}_{C_i}$ using Eq. (\ref{eq:lensemble});
\STATE Define a distribution on all possible subsets $\mathcal{P}_{\mathbb{L}}(Y_{C_i})$ using Eq. (\ref{eq:py});
\STATE Obtain a sample of $\mathcal{P}_{\mathbb{L}}(Y_{C_i})$;
\STATE Save nodes in the sample as $\overline{\mathcal{N}}(i)$.
\ENDFOR
\STATE \textbf{return} $\overline{\mathcal{N}}$
\end{algorithmic}
\end{algorithm}

\begin{algorithm}[!t]
\caption{D2GCN}
\label{alg:dgcn}
\textbf{Input}: A graph $G$ \\
\textbf{Output}: $x^{(L)}_i$ for all $i \in G$
\begin{algorithmic}[1] 
\STATE Build DFS path for all nodes;
\FOR{each level $l$}
\FOR{each node $i$ in $l$}
\STATE Find negative samples for $i$ using Algorithm \ref{alg:dfs-dpp};
\STATE Update node representation ${x_i}^{(l)}$ using Eq. (\ref{eq:dgcn});
\ENDFOR
\ENDFOR
\STATE \textbf{return} $x^{(L)}$
\end{algorithmic}
\end{algorithm}

\subsection{GCN Boosted by Diverse Negative Samples}

The classical GCN is only based on positive samples as shown in Eq. (\ref{eq:gcn}), which will inevitably lead to over-smoothing issues. With the negative samples from Algorithms  \ref{alg:dpp} or \ref{alg:dfs-dpp}, we propose the following new graph convolutional operation as 
\begin{equation}
\begin{aligned}
	{x_i}^{(l)} =& \sum_{j \in \mathcal{N}_i \cup \{i\}}\frac{1}{\sqrt{\text{deg}(i)}\cdot\sqrt{\text{deg}(j)}}\big(\Theta^{(l)} \cdot x_j^{(l-1)}\big)\\
	&- \omega \sum_{\bar{j} \in \overline{{\mathcal{N}}_i}}\frac{1}{\sqrt{\text{deg}(i)}\cdot\sqrt{\text{deg}(\bar{j})}}\big(\Theta^{(l)} \cdot x_{\bar{j}}^{(l-1)}\big)
\end{aligned}
\label{eq:dgcn}	
\end{equation} 
where $\overline{{\mathcal{N}}_i}$ is the negative samples of node $i$ and $\omega$ is a hyper-parameter to balance the contribution of the negative samples. It is also interesting to consider that all these negative samples form a virtual graph with the same nodes as before but with negative links between the nodes. When using message-passing framework for the mode learning, there are in fact two messages from each node: one is positive from neighbouring nodes and the other is negative from the negative samples. The positive messages push all the nodes with same semantics to have similar representations, while the negative messages push all nodes with different semantics to have different representations. This strategy is similar to clustering, where samples within same cluster are a small distance apart and large distances between samples indicate the sample is sitting in a different cluster. We believe that these negative messages are precisely what is missing from GCNs and a significant element in their advancement.

The final GCN boosted by diverse negative samples (D2GCN) with $L$ layers, is given in Algorithm \ref{alg:dgcn}. Excluding negative sampling, the computational cost is the double that of the original GCN. Note that, although GCNs are used as the base model here, these idea can be easily applied to other GNNs as well.

\section{Experiments}
\label{sec:experiments}

\subsection{Datasets}

The datasets we used are benchmark graph datasets in the literature: Citeseer, Cora and Pubmed \cite{sen2008collective}. The datasets include sparse bag-of-words feature vectors for each document as well as a list of document-to-document citation connections. Datasets are downloaded from Pytorch geometric\footnote{https://pytorch-geometric.readthedocs.io/en/latest/modules/\\datasets.html}. To better perform the experiments using DFS, we chose the maximum connected subgraph of each graph data. The datasets were split strictly in accordance with \cite{DBLP:conf/iclr/KipfW17}.


\subsection{Baselines}

\textbf{GCN} is the base model. We compare our sampling method with the only three negative sampling methods available to date, then put the selected samples into the convolution operation using Eq. (\ref{eq:dgcn}). The first method is to select negative samples in a purely random way, named \textbf{RGCN} \cite{kim2021find}. The second one is based on Monte Carlo chains, named \textbf{MCGCN} \cite{yang2020understanding}. The last one is based on  personalised PageRank, named \textbf{PGCN} \cite{ying2018graph}
To ensure the consistency and fairness of the experiments, all the graph convolution models had the same structure and were initialised and trained with the same methods. Note there is no other related negative sampling works in the field.

\subsection{Setup}

The experimental task was standard node classification. We set the length of DFS to 5 and the negative rate as a trainable parameter and trained all models with different numbers of layers in the range $\{2, \cdots, 6\}$ to test behaviour at increasing depths. Each model was trained for 200 epochs on Cora and Citeseer and for 100 epochs with Pubmed, using an Adam optimiser with a learning rate of 0.01. Tests for each model at each depth with each dataset were conducted 10 times. Moreover, all experiments were conducted on an Intel(R) Xeon(R) CPU @ 2.00GHz and NVIDIA Tesla T4 GPU. The code was implemented in PyTorch\footnote{The code is available at https://github.com/Wei9711/D2GCN.}.

\subsection{Metrics}

It was our goal to verify two capabilities of the proposed model: one is ability to improve the prediction performance and the other is to alleviate the over-smoothing problem. Hence, we used the following two metrics:

\begin{itemize}
\item \textbf{Accuracy} is the cross-entropy loss for the node label prediction on test nodes (the larger is the better);
\item \textbf{Mean Average Distance
(MAD)} \cite{chen2020measuring} reflects the smoothness of graph representation (the larger is the better):
\begin{equation}
\mathrm{MAD}=\frac{\sum_i D_i}{\sum_i 1\left(D_i\right)}, \quad D_i=\frac{\sum_j D_{i j}}{\sum_j 1\left(D_{i j}\right)}
\end{equation}

where $D_{i j}=1-\cos \left(x_i, x_j\right)$ is the cosine distance between the nodes $i$ and $j$.
\end{itemize}

\subsection{Results}

\begin{figure*}[!t]
\centerline{\includegraphics[width=\textwidth]{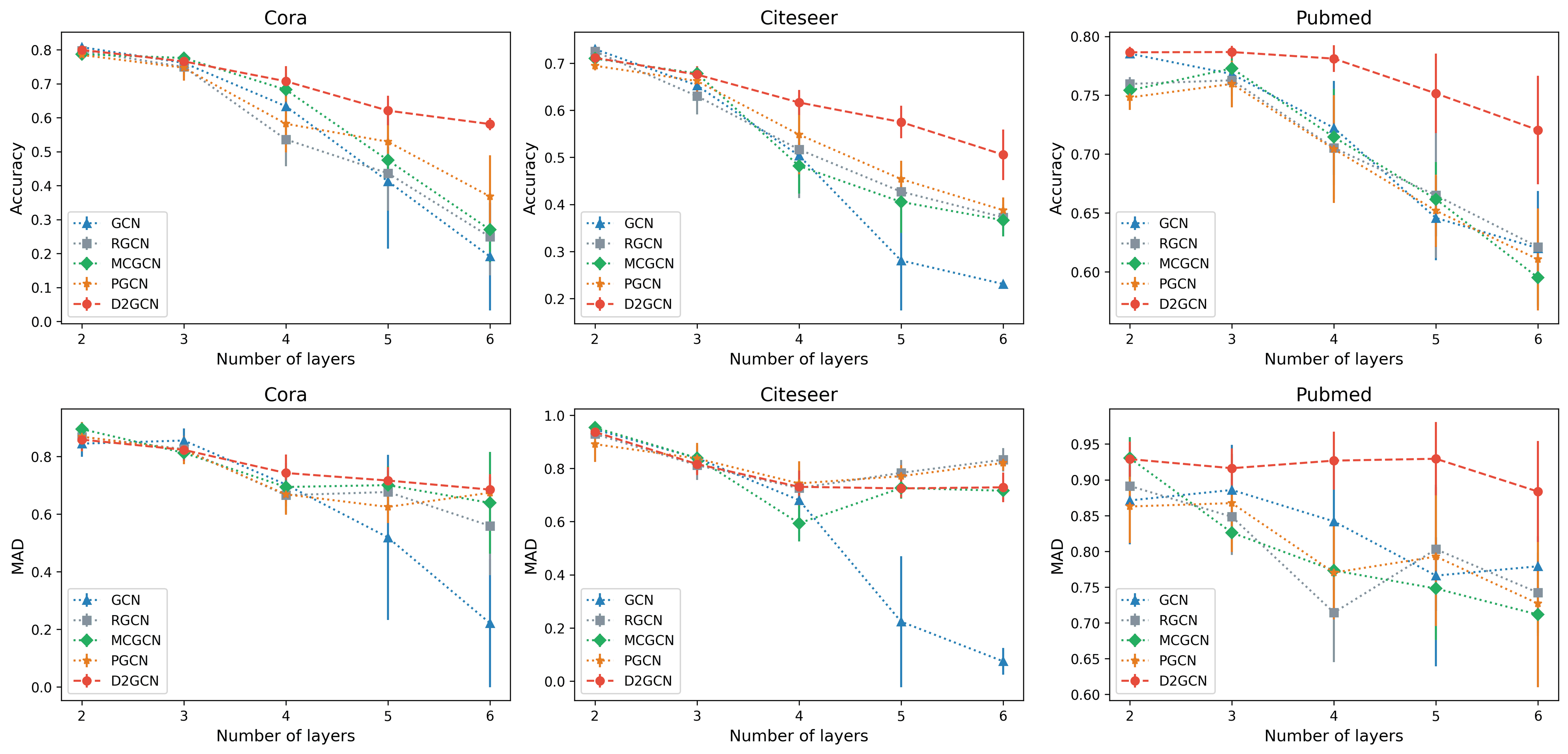}}
\caption{The Accuracy  and MAD of five models on three datasets with varying number of layers from 2 to 6. The x-axis denotes the layer number, and the mean and standard deviation of 10 runs are given for each model with each layer number.}
\label{fig:ACC+MAD}
\end{figure*}

\begin{figure*}[!t]
\centerline{\includegraphics[width=\textwidth]{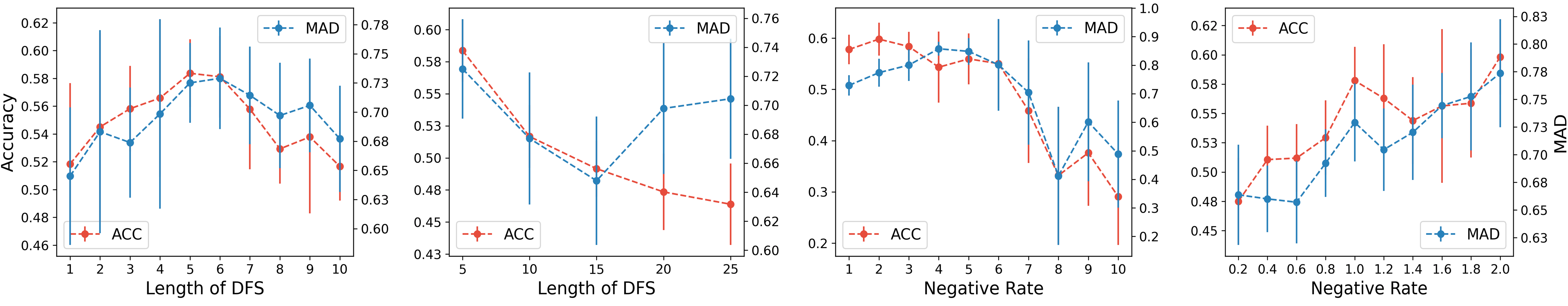}}
\caption{The Accuracy and MAD of D2GCN with 5 layers on Citeseer datasets in varying length of DFS and scale of negative rate. The mean and standard deviation of 10 runs are given for each setting.}
\label{fig:DFS-NR}
\end{figure*}

The results of five models with different numbers of layers on three datasets are shown in Fig. \ref{fig:ACC+MAD}. The performances of all five models on Cora and Citeseer were very similar in terms of both Accuracy and MAD at 2 and 3 layers. On Pubmed, our model was marginally better than the others. When the depth increased from 3 to 6, both the Accuracy and MAD of all models decreased to some extent. This is a consistent observation with the literature that increasing the depth of the network leads to a performance drop due to over-smoothing. We  also observed that the decreasing trends of RGCN, MCGCN and PGCN on Accuracy were very similar, while PGCN are slightly better than the other two on both Cora and Citeseer datasets, especially at layers 5 and 6. As for MAD, their trends were similar on Pubmed, but on the other two datasets GCN are much worse than the others. Moreover, although the MAD of RGCN, MCGCN and PGCN is close to our D2GCN on Cora and Citeseer, especially at layers 5 and 6 on Citeseer RGCN and PGCN even surpass our method, they are much less accurate than D2GCN. This observation suggests that adding negative samples to the convolution does reduce the over-smoothing to some extent, but choosing the appropriate negative samples is is not trivial. Additional effort needs to be given to the procedure for selecting negative samples. 

As shown in the Fig. \ref{fig:ACC+MAD}, our proposed model achieved consistently the best performance in terms of accuracy on all datasets. For MAD, it also performs outstandingly in particular on the Cora and Pubmed datasets. At first, we observed that the performance of our model also decreased along with the increasing depth. However, the rate at which performance decreased was much slower than for the other two models – and almost flat on Pubmed. Secondly, in terms of MAD, the performance of our model generally exceeds the others, especially on Pubmed by a large margin. These observations verify that our idea is able to significantly alleviate the over-smoothing problem and in turn improve the prediction accuracy. As a final observation, we found that the variance of our model was smaller than the other four. One possible reason is that the diverse negative samples may quickly change the current node representation rather than keep sticking around the initialisation when only positive samples are used. 

The sensitivity of hyper-parameters is analysed and shown in Fig. \ref{fig:DFS-NR} where we trained 5-layer D2GCN on Citeseer dataset in varying length of DFS and scale of negative rate. We observe that as negative rate or DFS-length goes up, the trends of both Accuracy and MAD are roughly the same, increasing first and then decreasing. For the length of DFS, the outstanding results are obtained when it is equal to 5 or 6. For the negative rate, it achieves the same performance as the trainable parameters, when it is equal to $\{1,2,3\}$. 

We further show results on three different types of datasets in Tab.\ref{new_result}: protein graph, gamer graph, paper graph. We compared two SOTA methods and two latest methods for over-smoothing include: GATv2 \cite{brody2021attentive} , GraphSage \cite{hamilton2017inductive} and MAD \cite{chen2020measuring} and DGN \cite{DBLP:conf/nips/Zhou0LZCH20}. Setting is same with the above and all models are with 4 layers. Our method has consistently achieved  better performance.

\begin{table}[!t]
    \centering
    \begin{tabular}{c|ccc}
    \hline
        \textbf{Method} 
        & \textbf{Proteins} 
        & \textbf{Twitch(EN)} 
        & \textbf{ogbn-arxiv} 
        \\ \hline
        D2GCN & \textbf{.74} $\pm$ \textbf{.02} & \textbf{.58} $\pm$ \textbf{.01} & \textbf{.66} $\pm$ \textbf{.01} \\ 
        GATv2 & .66 $\pm$ .02 & .54 $\pm$ .03 & .31 $\pm$ .06\\ 
        GraphSAGE& .66 $\pm$ .02 & .55 $\pm$ .01 & .58 $\pm$ .02 \\ 
        MAD & .66 $\pm$ .02 & .55 $\pm$ .02 & * \\
        DGN & .64 $\pm$ .01 & .52 $\pm$ .01 & .60 $\pm$ .01 \\ \hline
    \end{tabular}
\footnotesize{$*$ MAD on obgn-arxiv didn't finish one run after more than 12h.}
\caption{Accuracy on other three different types of datasets.}
\label{new_result}
\end{table}

\section{Conclusion and Future Work}
\label{sec:conclusion}

In this paper, we identified the importance of negative samples to GCNs and described the criteria for what constitutes a good negative sample. We introduced DPP to select meaningful negative samples from the dark world of the whole graph. To the best of our knowledge, we are the first to introduce DPP to GCNs for negative sampling and the first to fuse the negative samples into the graph convolution. We further presented a DFS-based heuristic approximation method to greatly reduce the computational cost. The experimental evaluations shows that the proposed D2GCN consistently delivers better performance than alternative methods. In addition to greater predictive accuracy, the method also helps to prevent over-smoothing. With this study, we identify that negative samples are important to graph neural networks and should be considered in future works. Note that the proposed idea can be applied to other graph neural networks apart from GCN.

In future research, we will continue to investigate how to speed-up the DPP sampling process in the algorithm. Another interesting follow-up work would be to investigate more effective aggregation of positive and negative samples as the current solution may lose some information when summing the samples together – especially when the samples are diverse.

\section*{Acknowledgements}
 This work is supported by the Australian Research Council under Australian Laureate Fellowships FL190100149 and Discovery Early Career Researcher Award DE200100245.

\bibliography{aaai22} 

\end{document}